\documentclass[runningheads]{llncs}

\usepackage{eccv}

\usepackage{eccvabbrv}
\usepackage{graphicx}
\usepackage{booktabs}
\usepackage{amsmath,amssymb,amsfonts}
\usepackage{algorithmic}
\usepackage{algorithm}
\usepackage{textcomp}
\usepackage{xcolor}
\usepackage{url}
\usepackage{siunitx}
\usepackage{multirow}
\usepackage{verbatim}
\usepackage{bm}
\usepackage[accsupp]{axessibility}

\usepackage{hyperref}
\usepackage{orcidlink}

\begin{document}

\title{PRISM: Predictive Representation of Interaction Style and Motion for Social Robot Navigation}
\titlerunning{PRISM for Social Robot Navigation}

\author{Bo-Han Chen\inst{1} \and Hiromu Taketsugu\inst{2} \and Norimichi Ukita\inst{2}}
\authorrunning{B.-H. Chen et al.}

\institute{National Chung Hsing University \and Toyota Technological Institute}

\maketitle

\begin{abstract}
Humans often observe others before interacting and adjust their behavior accordingly. Robot navigation in crowds, however, often represents pedestrians mainly by observed geometric states, leaving individual differences in interaction tendencies implicit. We propose PRISM (Predictive Representation of Interaction Style and Motion), a framework that infers interaction traits from passive observations of human-human interactions. PRISM encodes human trajectories into a continuous ordinal latent space with a transformer encoder trained by Rank-N-Contrast loss, and pairs each inferred trait with a temporal-stability score supplied to the navigation policy. In randomized crowd simulations, PRISM reduces collision rates over the geometry-only baseline and yields small improvements in navigation-time and path-length metrics. These results suggest the utility of passive latent-trait inference for social navigation in dynamic crowds.
\end{abstract}
%-------------------------------------------------------------------------

\begin{figure}[t]
    \centering
    \includegraphics[width=0.85\linewidth]{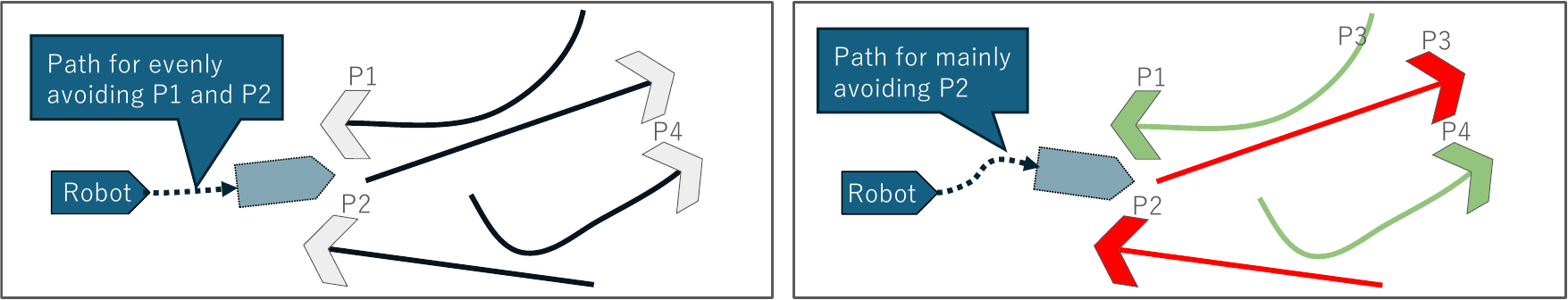}\\
    (a) No personalization \hspace*{24mm} (b) PRISM ~~~~~
    \caption{Overview. Colored pedestrian trajectories indicate observed interaction histories used to infer heterogeneous interaction styles, including unyielding and yielding behavior shown in red and green, respectively. (a) Without personalization, the robot relies on geometric states and may delay avoidance. (b) PRISM augments the navigation policy with inferred Interaction Style Vectors and confidence scores, producing a robot trajectory that better anticipates different yielding tendencies.}
    \label{fig:teaser}
    \vspace*{-1mm}
\end{figure}

\section{Introduction}
\label{section:introduction}

As mobile robots move into human-centric environments, social acceptability becomes as important as efficiency~\cite{kruse2013humanaware,moller2021survey,singamaneni2024survey}. A key difficulty is \textit{latent behavioral heterogeneity}~\cite{sih2004behavioral}: individuals differ in how strongly they yield, keep distance, or push forward in shared spaces.
%%%
Current socially-aware navigation frameworks often represent pedestrians mainly through their trajectories represented by positions and velocities~\cite{liu2024height,chen2017socially,chen2019crowd}. Such trajectories obtained by multi-object tracking~\cite{DBLP:journals/cviu/UkitaO16,Shim_2025_CVPR,SAMIDARE} can contain useful cues, although individual yielding or assertive tendencies remain entangled with geometric motion. Active probing can identify interaction parameters by disturbing an agent~\cite{pandya2022safe}, yet this violates the non-disturbance principle desirable in human-robot interaction.

We instead follow \textit{social eavesdropping}~\cite{nowak2005evolution,johnstone2001eavesdropping,tibbetts2020wasps}, in which a robot passively observes human-human interactions and infers an agent's \textit{Interaction Style} from observed interaction histories without perturbing that agent. We propose \textbf{PRISM} (Predictive Representation of Interaction Style and Motion), which learns a continuous ordinal representation of interaction styles and provides it to a navigation policy. Figure~\ref{fig:teaser} illustrates this passive observation setting.

PRISM is designed for interactive agents in dynamic crowds, and encodes relative motion during encounters as continuous state variables, without probing pedestrians or requesting labels.

Our contributions are as follows:
(1) a passive mechanism for inferring behavioral traits from trajectories using Rank-N-Contrast loss,
(2) a temporal confidence estimate that is provided to the policy alongside the trait, and
%a confidence-aware gating module that reduces the influence of uncertain trait estimates, and
(3) an evaluation in heterogeneous crowd simulations showing improved collision-based safety metrics when our interaction-style representation is integrated into robot navigation.

%%%%%%%%%%%%%%%%%%%%%%%%%%%%%%%%%%%%%%%%%%%%%%%%%%%%%%%%%%%%%%%%%%%%%%

\section{Related Work}
\label{section:related}

Early methods such as the Social Force Model~\cite{helbing1995social} and Reciprocal Velocity Obstacles~\cite{vanDenBerg2008rvo} use fixed interaction rules, whereas crowd-aware attention models~\cite{chen2019crowd,chen2017socially}, intention-aware interaction graphs~\cite{liu2022intentionaware}, and the Heterogeneous Interaction Graph Transformer (HEIGHT)~\cite{liu2024height} learn richer social dynamics. These methods use observed pedestrian trajectories for navigation and forecasting without exposing per-person latent traits to the policy.

Trajectory and motion prediction, such as Social-LSTM~\cite{7780479}, Social-GAN~\cite{gupta2018socialgan}, and related models~\cite{lee2017desire,kosaraju2019socialbigat,salzmann2020trajectron++,mangalam2021pecnet,DBLP:conf/iccv/0001U23,DBLP:conf/cvpr/TaketsuguO0NU25,DBLP:journals/tmlr/MaedaCUK26}, captures social interactions for anticipating pedestrian motion. Recent studies~\cite{kothari2022motionstyle,liu2022causalmotion} address style variation and distribution shift. These works target future-motion prediction, whereas PRISM infers an interaction-style variable.

Personalized prediction methods such as T4P~\cite{Park2024T4P}, Interactive Adjustment~\cite{Sun2025InteractiveAdjustment}, DisDis~\cite{Chen2021DisDis}, meta-learning-based prediction~\cite{Zhu2022MetaLearningTrajectory}, and MemoNet~\cite{Xu_2022_CVPR} adapt prediction to individual histories or memory. PRISM differs by using interaction history for personalization and learning a temporally consistent interaction-style representation for downstream planning. The goal is not to improve open-loop trajectory forecasting, but to provide a compact trait variable that a closed-loop navigation policy can use while observing agents.

%%%%%%%%%%%%%%%%%%%%%%%%%%%%%%%%%%%%%%%%%%%%%%%%%%%%%%%%%%%%%%%%%%%%%%%%%%%%%%%%

\begin{figure}[t]
    \centering
    \includegraphics[width=\linewidth]{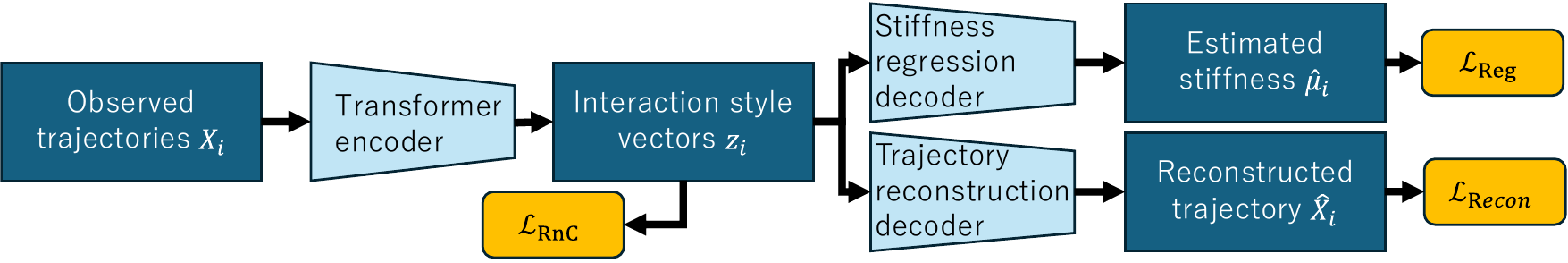}
    \caption{PRISM architecture. A transformer encoder maps trajectories $\bm{X}_i$ to an Interaction Style Vector $\bm{z}_i$, which is structured by RnC loss and decoded for stiffness regression and input reconstruction.}
    \label{fig:prism_architecture}
\end{figure}

\section{Proposed Method}

\subsection{PRISM}

Figure~\ref{fig:prism_architecture} shows the PRISM architecture. Given agents
$\mathcal{S}_t=\{P_0,P_1,\dots,P_K\}$, $P_0$ denotes the ego-agent, i.e., the
robot, and $P_1,\dots,P_K$ denote neighboring agents. For each $P_i$, PRISM constructs a trajectory history
$\bm{X}_i\in\mathbb{R}^{T\times4}$ with $T=30$ frames, corresponding to
3.0 seconds at 10 frames per second. Its feature at time step $\tau$ is
$
\bm{x}_{i,\tau}
=
[p^x_{i,\tau},p^y_{i,\tau},v^x_{i,\tau},v^y_{i,\tau}]^\top,
$
where $(p^x_{i,\tau},p^y_{i,\tau})$ and
$(v^x_{i,\tau},v^y_{i,\tau})$ denote the position and velocity of $P_i$,
respectively. The trajectories are represented in a local coordinate frame
centered at the robot's initial position and aligned with its initial velocity,
providing invariance to global translation and rotation.

For a target neighbor $P_i$, the interaction context
$\mathcal{C}_i=\mathcal{S}_t\setminus\{P_i\}$ consists of the robot and the
other neighbors. A dual-stage Transformer encoder~\cite{vaswani2017attention}
first applies temporal self-attention to the target trajectory and then uses
cross-attention to incorporate the context trajectories. Invalid and padded
context tokens are masked. Standard multi-head attention, positional encoding,
residual connections, layer normalization, and feed-forward layers are used.
The resulting contextualized features are average-pooled over time and
projected by a two-layer multilayer perceptron with a GELU activation to obtain
the Interaction Style Vector $\bm{z}_i\in\mathbb{R}^{D}$, where $D=128$.

PRISM uses relative motion during encounters, including how agents approach,
change velocity, and adjust their paths, as evidence of interaction style.
It requires only the position and velocity histories available to the navigation system, without semantic interaction labels. PRISM processes the most recent $T = 30$ frames and updates the interaction-style estimate every ten control steps, holding it constant in between. Passive
observation means that the robot does not perturb pedestrians or request behavior
labels.
%; no pre-navigation observation phase is required.

\subsection{Rank-N-Contrast Supervision}

In simulation, each agent has a social stiffness label $\mu\in[0,1]$, where low values indicate aggressive, weakly yielding behavior, and high values indicate cautious, strongly yielding behavior. A binary contrastive loss would require discretizing this trait, incorrectly separating similar agents near an arbitrary threshold. We use Rank-N-Contrast (RnC) loss~\cite{zha2020rank} to encourage ordinal geometry.

For an anchor $i$ in a mini-batch $\mathcal{B}$, all other samples are sorted by label distance $|\mu_i-\mu_k|$. Let $\Omega_i=\{k_1,k_2,\dots\}$ be this sorted index set, ordered such that $|\mu_i-\mu_{k_1}|\leq|\mu_i-\mu_{k_2}|\leq\cdots$. The RnC loss is
\begin{equation}
    \mathcal{L}_{\text{RnC}} = - \sum_{i \in \mathcal{B}} \sum_{m=1}^{|\Omega_i|} \log \frac{\exp(-\|\bm{z}_i - \bm{z}_{k_m}\|_2 / \tau_{\mathrm{RnC}})}{\sum_{j=m}^{|\Omega_i|} \exp(-\|\bm{z}_i - \bm{z}_{k_j}\|_2 / \tau_{\mathrm{RnC}})},
\end{equation}
where $\tau_{\mathrm{RnC}}=2.0$ denotes the temperature parameter. This loss penalizes cases in which a sample with a dissimilar stiffness score is embedded closer to the anchor than a sample with a similar score. It does not guarantee manifold smoothness. It provides a rank-based training signal that encourages nearby stiffness values to have nearby embeddings and distant values to be separated. This structure is useful for reinforcement learning because the policy receives $\bm{z}_i$ as part of its state.

The encoder is trained with reconstruction and regression objectives:
\begin{align}
    \mathcal{L}_{\text{Recon}} &= || \hat{\bm{X}}_i - \bm{X}_i ||_2^2, \\
    \mathcal{L}_{\text{Reg}} &= || \hat{\mu}_i - \mu_i ||_2^2, \\
    \mathcal{L}_{\text{Total}} &= \lambda_{\text{Recon}} \mathcal{L}_{\text{Recon}} + \lambda_{\text{RnC}} \mathcal{L}_{\text{RnC}} + \lambda_{\text{Reg}} \mathcal{L}_{\text{Reg}},
\end{align}
where $\hat{\bm{X}}_i$ denotes the reconstructed trajectory, $\hat{\mu}_i$ denotes the regression-head stiffness prediction, and $\lambda_{\text{Recon}}=1.0$, $\lambda_{\text{RnC}}=1.0$, and $\lambda_{\text{Reg}}=0.1$.
The regression head is used only during representation learning. During policy learning, the planner receives a projection of the latent vector and
the confidence score defined later in Eq.~(\ref{eq:alpha}), rather than the
ground-truth simulator label, so the closed-loop policy does not receive
privileged stiffness labels during inference.

\subsection{Temporal Uncertainty Estimation}

To quantify the stability of its trait estimates, PRISM tracks embedding volatility with an exponentially weighted moving variance:
\begin{equation}
    \sigma^2_t = (1 - \beta)\sigma^2_{t-1} + \beta \|\hat{\bm{z}}_t - \bm{z}_{t-1}\|_2^2,
\end{equation}
where $\hat{\bm{z}}_t$ denotes the instantaneous encoder output, $\bm{z}_{t-1}$ denotes the previous smoothed estimate, and $\beta=0.1$ denotes the smoothing factor. This recursive estimator maintains a fading memory of embedding stability while emphasizing recent observations. The resulting volatility is mapped to a confidence score $\alpha_t\in[0,1]$ using a radial basis function (RBF) kernel:
\begin{equation}
%    \alpha_t = \exp\left(-\frac{\sigma^2_t}{2\gamma^2}\right),
    \alpha_t = \exp\left(- \sigma^2_t / (2\gamma^2) \right),
    \label{eq:alpha}
\end{equation}
where $\gamma=1.0$ denotes the RBF kernel width.
A high volatility produces a small $\alpha_t$, indicating that the style
estimate is currently unstable. The planner receives the interaction-style representation together with $\alpha_t$, allowing the navigation policy to account for its temporal stability. Before $T=30$ frames of history are available, the policy receives a zero vector with $\alpha_t=0$. The estimate is also held fixed while no neighbor is within $2.5$ m, so that frames without a nearby agent do not perturb it.

%%%%%%%%%%%%%%%%%%%%%%%%%%%%%%%%%%%%%%%%%%%%%%%%%%%%%%%%%%%%%%%%%%%%%%%%%%%%%%%%

\begin{table}[t]
    \centering
    \caption{Phase 1 ablation study on latent representation.}
    \vspace*{-2mm}
    \resizebox{\columnwidth}{!}{%
    \begin{tabular}{lccc}
        \toprule
        \textbf{Configuration} & \textbf{Stiffness MAE} $\downarrow$ & \textbf{Manifold Rank ($\rho$)} $\uparrow$ & \textbf{Recon. ADE (m)} $\downarrow$ \\
        \midrule
        Recon Only & 0.248 & 0.019 & \textbf{0.006} \\
        Reg Only & 0.214 & 0.038 & 0.738 \\
        RnC Only & 0.247 & 0.162 & 0.858 \\
        Recon + Reg & 0.207 & 0.043 & 0.011 \\
        RnC + Reg & 0.219 & \textbf{0.189} & 0.771 \\
        Recon + RnC & 0.247 & 0.166 & 0.028 \\
        \midrule
        \textbf{Full Model (PRISM)} & \textbf{0.203} & 0.183 & 0.027 \\
        \bottomrule
    \end{tabular}
    }
    \label{tab:ablation}
    \vspace*{-2mm}
\end{table}

\section{Experiments}

\subsection{Simulation Environment and Crowd Heterogeneity}

We evaluate PRISM in a custom PyBullet simulator based on HEIGHT~\cite{liu2024height}. The arena is a $9m\times9m$ square with 8--12 randomly placed static obstacles modeled as furniture. The robot is a Turtlebot with non-holonomic kinematics and maximum linear velocity $v_{\max}=0.5$ m/s. It observes the environment through simulated two-dimensional light detection and ranging (LiDAR) with a $360^\circ$ field of view (FOV) and a 25m range. This observation is used by the navigation policy only. PRISM instead consumes ground-truth pedestrian positions with no observation noise, and velocities are finite differences of consecutive positions. Neighbor identity is provided by the simulator, so detection and tracking errors are outside the scope of this study.
%%%
Dynamic pedestrians are governed by an
%modified
Optimal Reciprocal Collision Avoidance (ORCA) policy~\cite{vanDenBerg2010orca}. ORCA provides controllable interactive pedestrian motion and interpretable parameters. Each agent has a latent social stiffness $\mu_i\in[0,1]$ that linearly interpolates its navigation parameters:
%\begin{equation}
%    \Phi_i = \mu_i \Phi_{\text{cautious}} + (1 - \mu_i) \Phi_{\text{aggressive}}.
%\end{equation}
$\Phi_i = \mu_i \Phi_{\text{cautious}} + (1 - \mu_i) \Phi_{\text{aggressive}}$.
Here, $\Phi_i$ comprises the ORCA time horizon, safety buffer, and neighbor-detection range. Cautious agents ($\mu=1$) use a long time horizon ($T_{\mathrm{hor}}=10.0$s), a safety buffer ($0.15$m), and a far detection range (10.0m). Aggressive agents ($\mu=0$) use a short time horizon ($T_{\mathrm{hor}}=2.0$s), zero safety buffer, and a 2.0m detection range.
Note that $\mu$ is a controllable simulation attribute, not a directly observed label in real crowds.

\subsection{Training and Evaluation}

Training has two phases. First, the PRISM encoder is trained on $10^5$ interaction frames using ground-truth $\mu$ for the RnC ordering signal. The regression and reconstruction heads are used only during this representation-learning phase. Second, the encoder is frozen and integrated into the HEIGHT navigation policy. The HEIGHT architecture and reward design follow the original paper~\cite{liu2024height}.

For each neighbor $P_i$, PRISM augments the original HEIGHT geometric feature $\bm{h}^{\mathrm{geom}}_{i,t}$ with the inferred style vector and confidence score:
$\bm{h}^{\mathrm{PRISM}}_{i,t} = [\bm{h}^{\mathrm{geom}}_{i,t};\tilde{\bm{z}}_{i,t};\alpha_{i,t}]$, where $\tilde{\bm{z}}_{i,t}\in\mathbb{R}^{8}$ is a linear projection of $\bm{z}_{i,t}$, so each neighbor contributes nine additional values.
While $\bm{h}^{\mathrm{PRISM}}_{i,t}$ is augmented, the remaining HEIGHT architecture is unchanged. The frozen PRISM encoder is evaluated every 10 control steps, and policy-learning gradients are not propagated into it.

We evaluate 500 randomized episodes per setting. Success rate is the ratio of episodes in which the robot reaches its goal without collision. The overall collision rate is the ratio of episodes ending in collision, with human and obstacle collisions reported separately. The timeout rate is the ratio of episodes that neither reach the goal nor collide within the maximum duration.
%Navigation time and path length are averaged over successful episodes.
%%%Standard HEIGHT serves as the baseline and receives only $\bm{h}^{\mathrm{geom}}_{i,t}$; PRISM receives $\bm{h}^{\mathrm{PRISM}}_{i,t}$.

\subsection{Results: Latent Representation}

\begin{figure}[t]
    \centering
    % Row 1
    \begin{minipage}{0.235\textwidth}
        \centering \includegraphics[width=\linewidth]{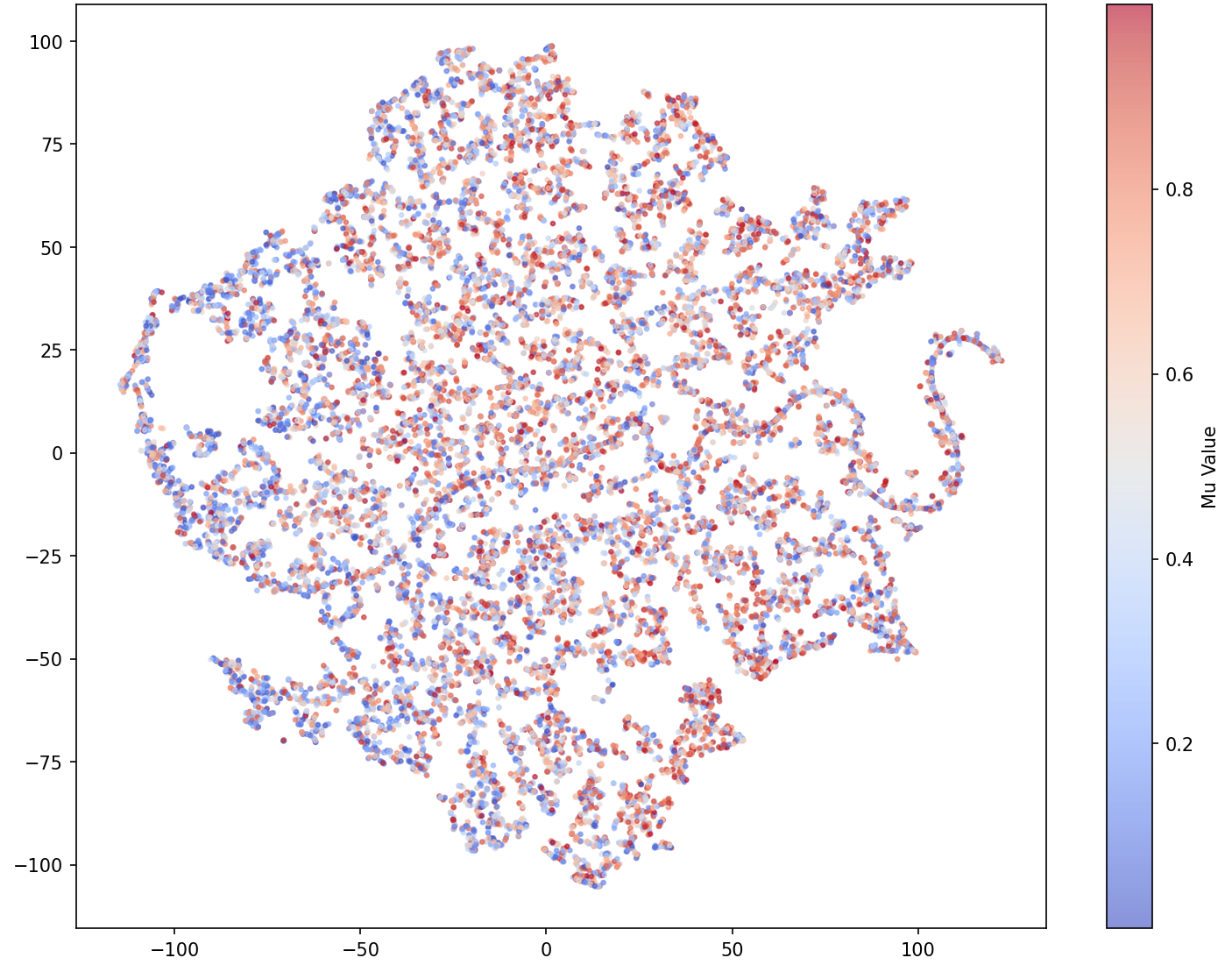}
        \centerline{(a) T4P}
    \end{minipage}\hfill
    \begin{minipage}{0.235\textwidth}
        \centering \includegraphics[width=\linewidth]{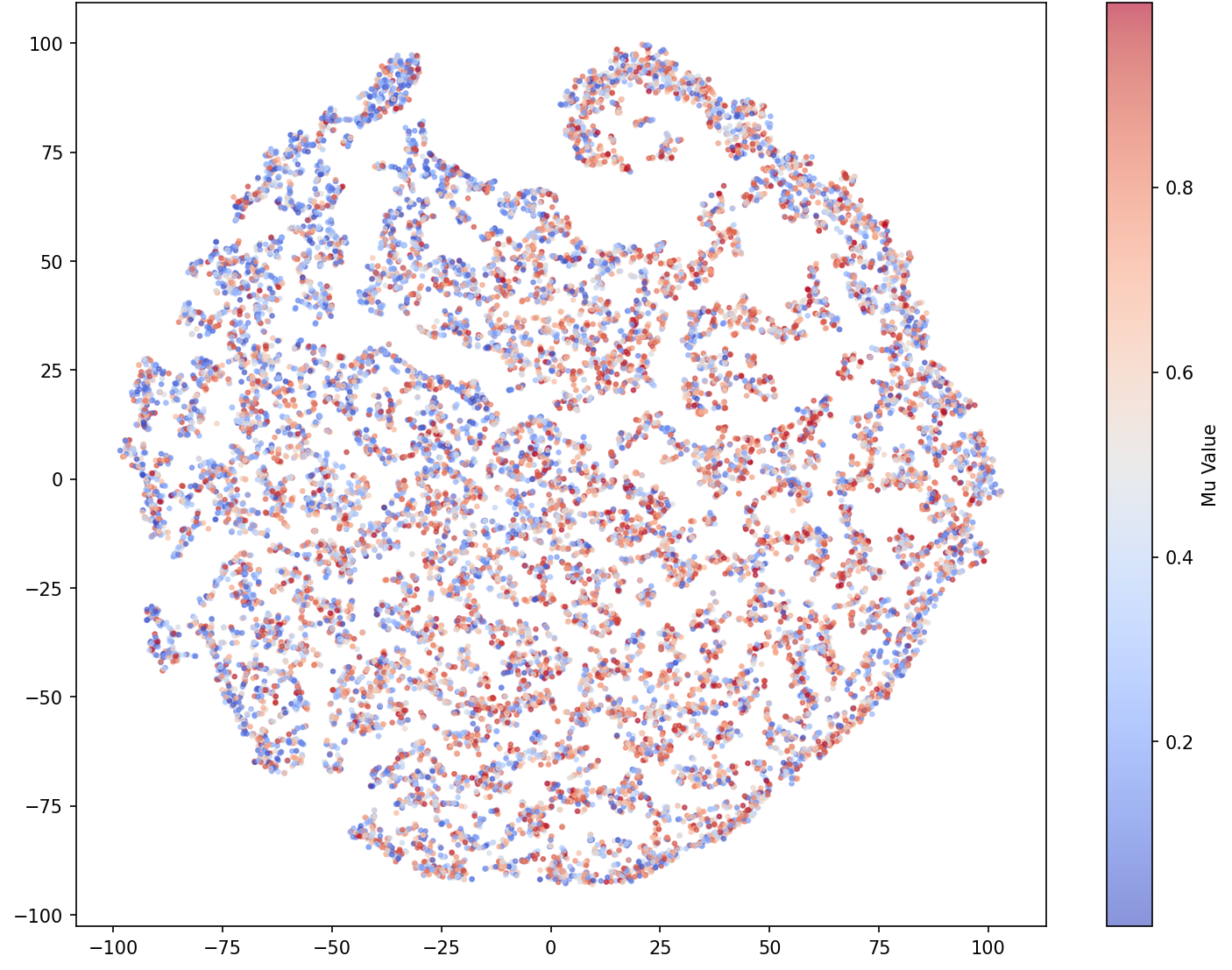}
        \centerline{(b) Recon Only}
    \end{minipage}\hfill
    \begin{minipage}{0.235\textwidth}
        \centering \includegraphics[width=\linewidth]{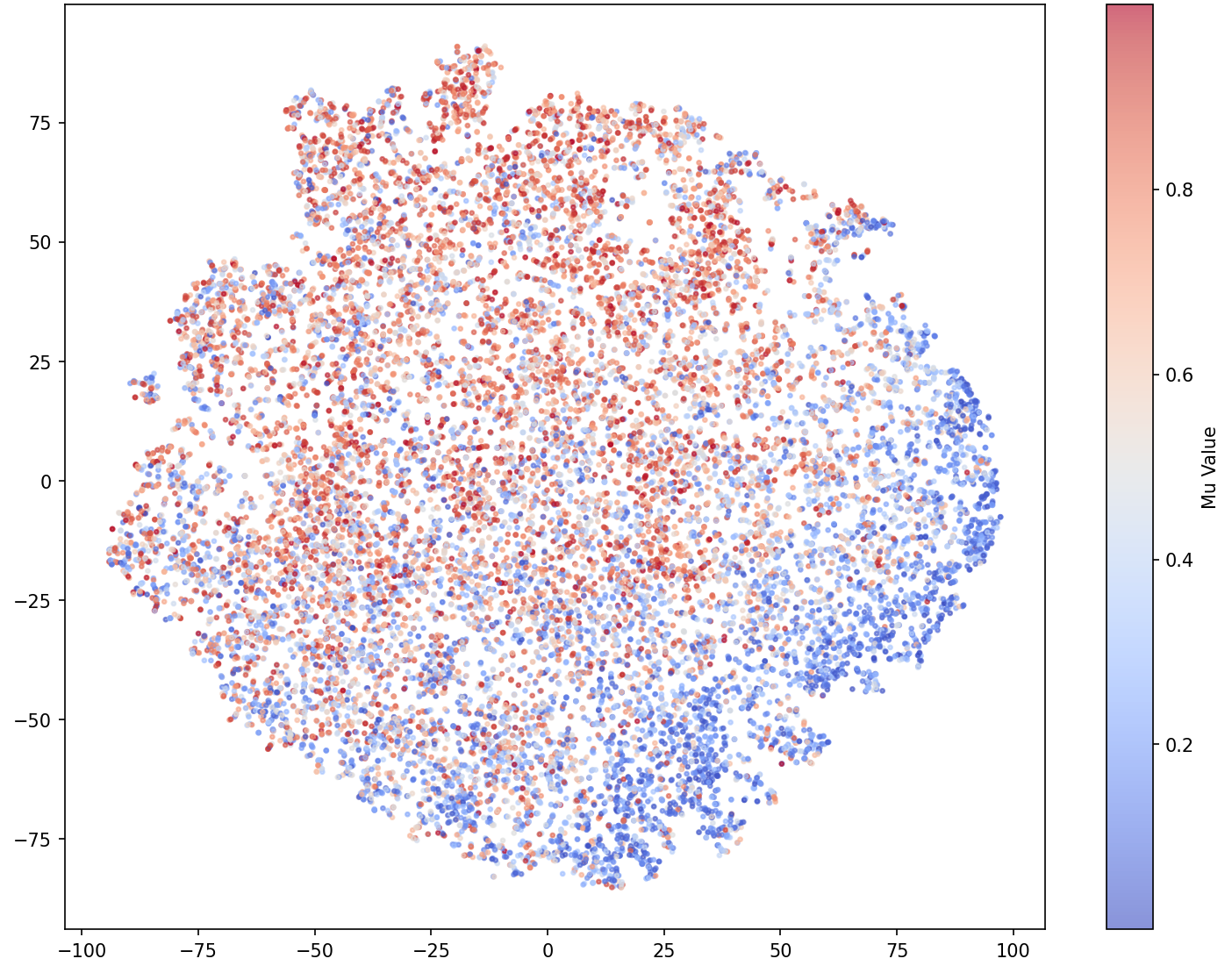}
        \centerline{(c) Reg Only}
    \end{minipage}\hfill
    \begin{minipage}{0.235\textwidth}
        \centering \includegraphics[width=\linewidth]{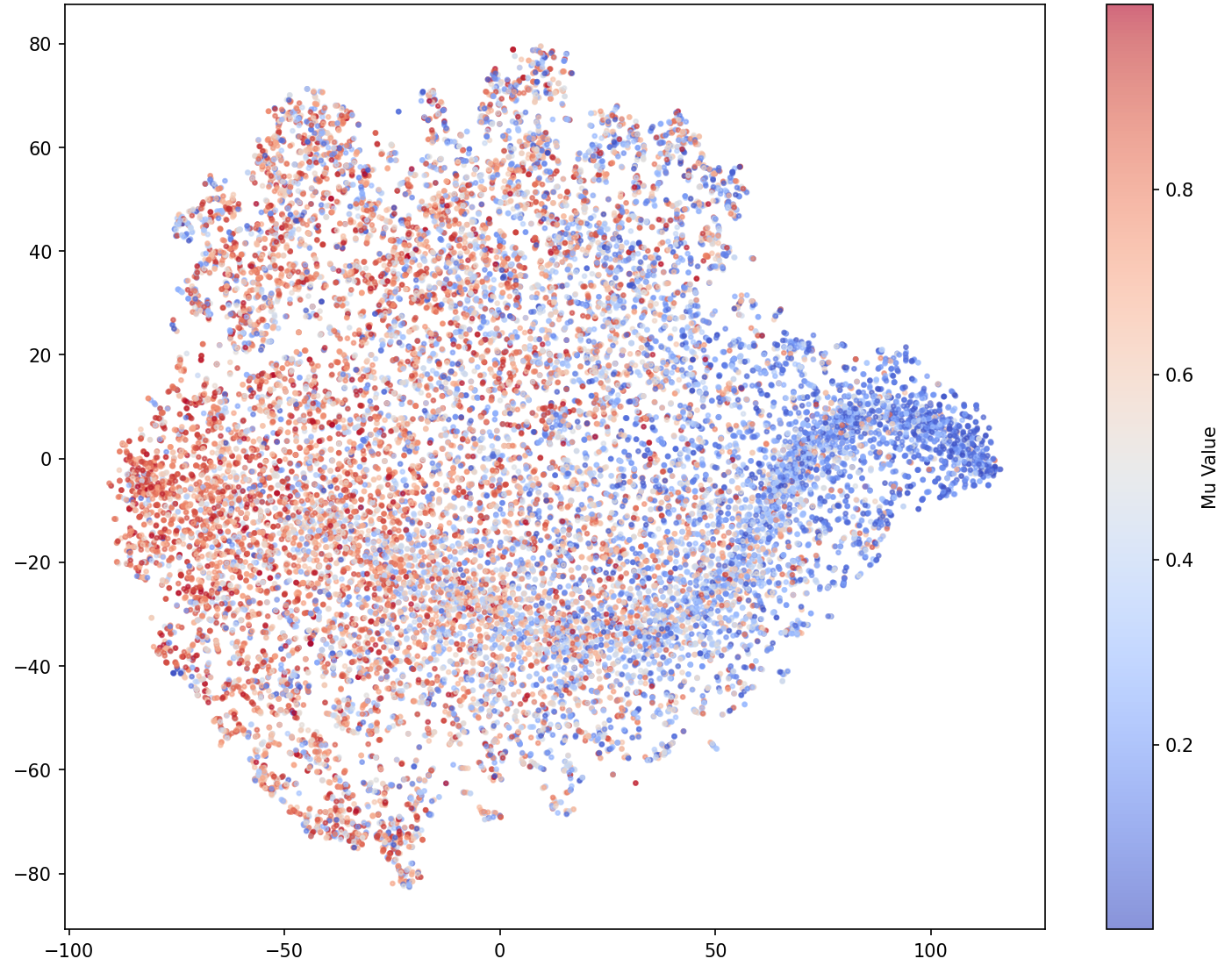}
        \centerline{(d) RnC Only}
    \end{minipage}
    
    \vspace{0.05cm}
    % Row 2
    \begin{minipage}{0.235\textwidth}
        \centering \includegraphics[width=\linewidth]{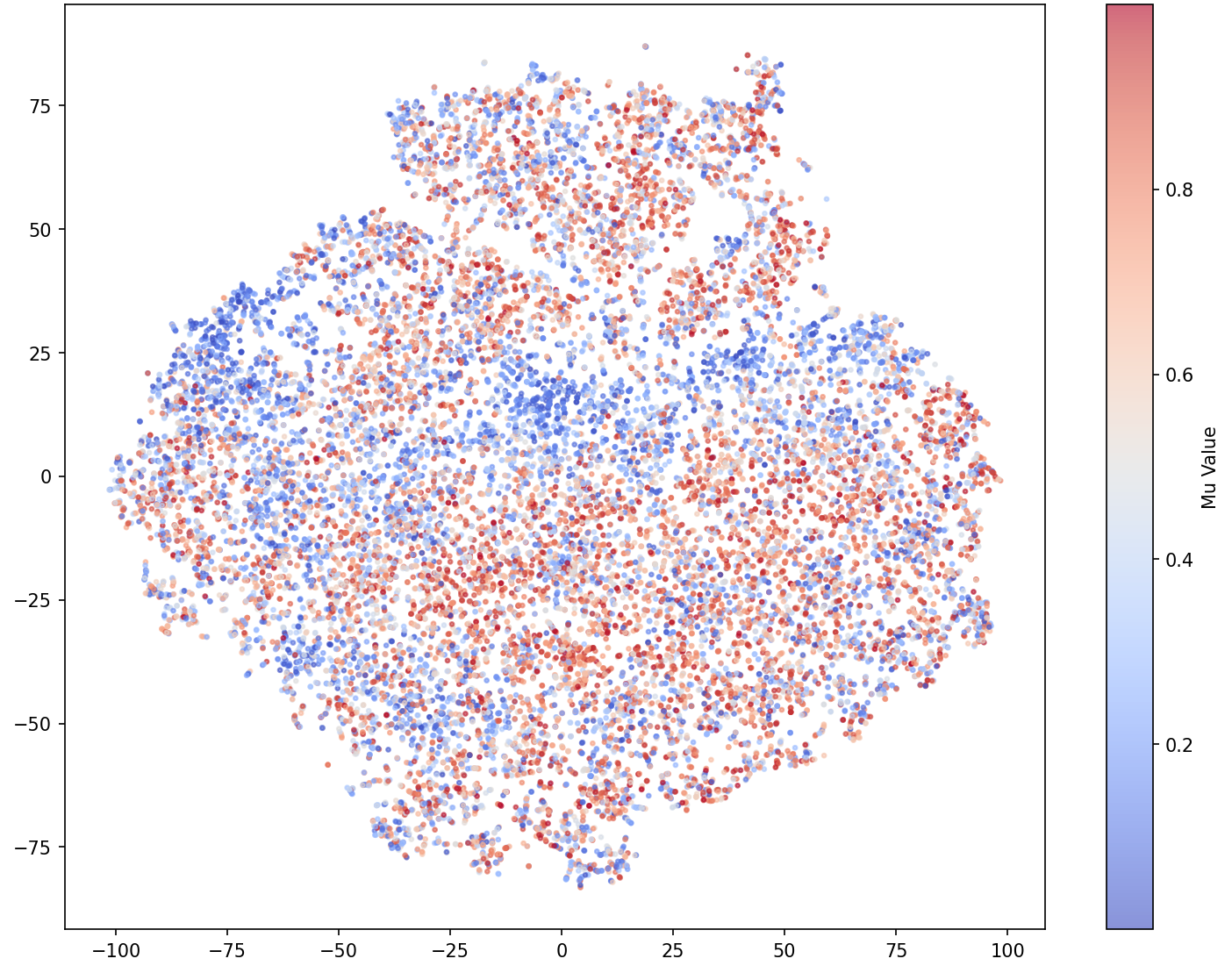}
        \centerline{(e) Recon + Reg}
    \end{minipage}\hfill
    \begin{minipage}{0.235\textwidth}
        \centering \includegraphics[width=\linewidth]{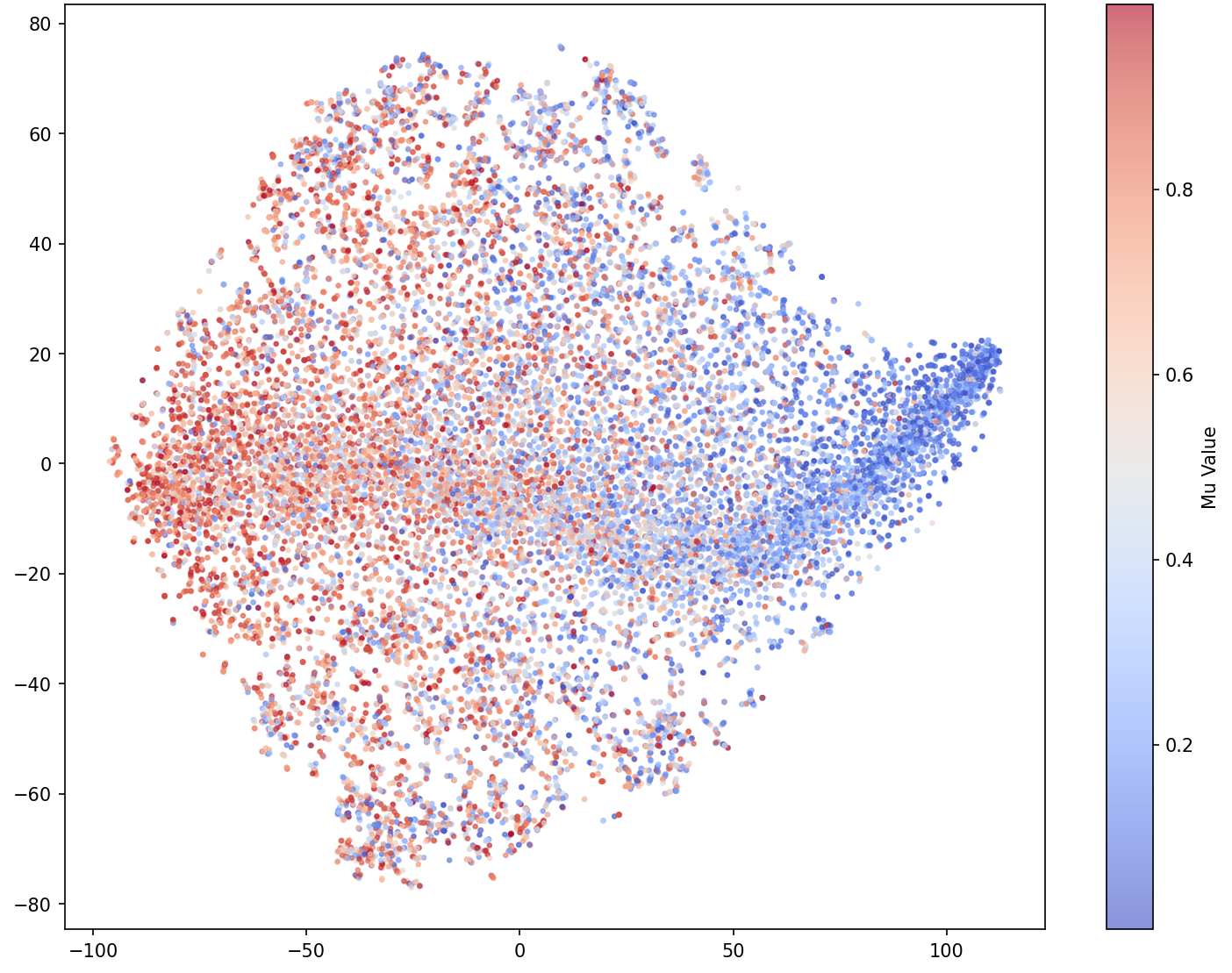}
        \centerline{(f) RnC + Reg}
    \end{minipage}\hfill
    \begin{minipage}{0.235\textwidth}
        \centering \includegraphics[width=\linewidth]{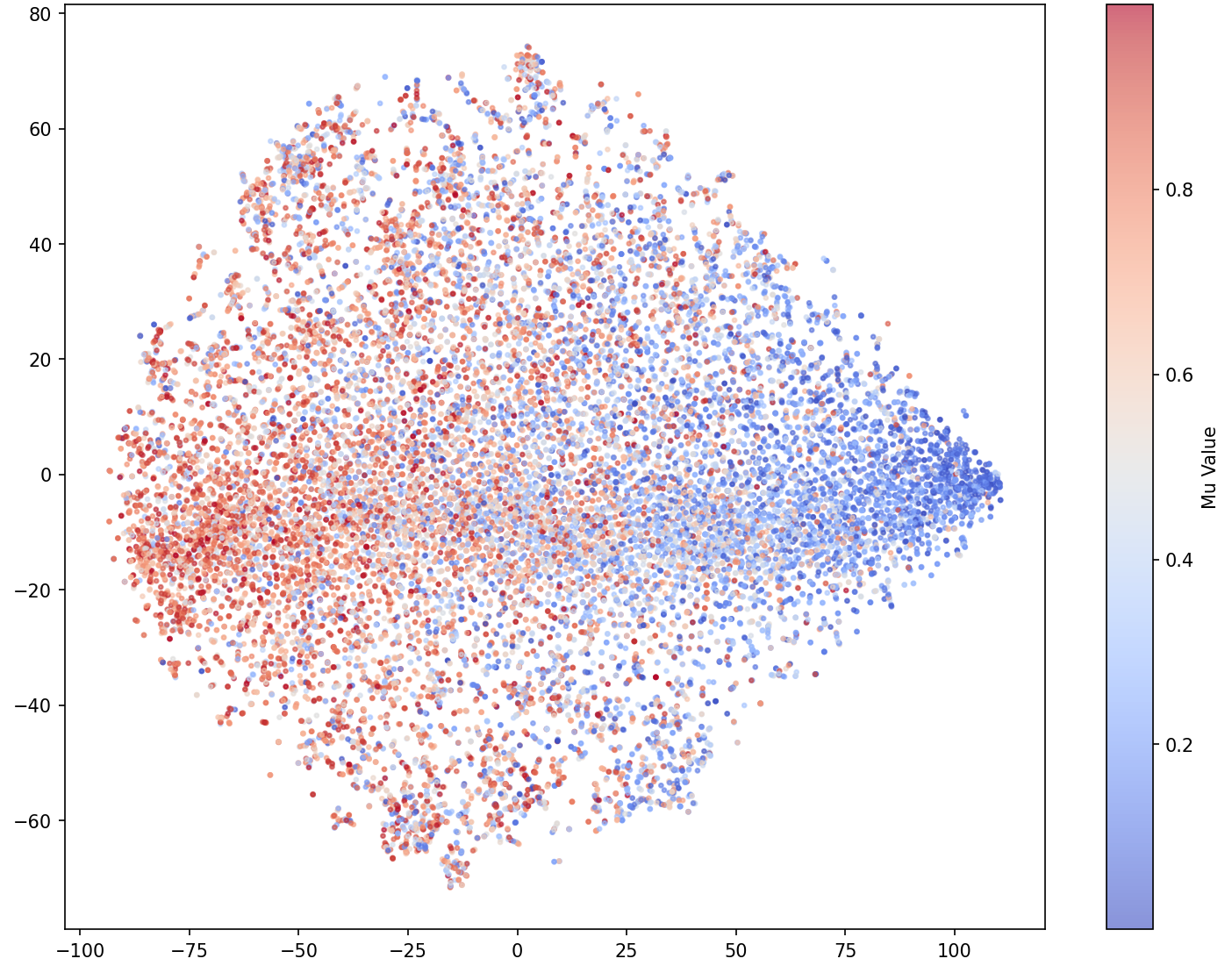}
        \centerline{(g) Recon + RnC}
    \end{minipage}\hfill
    \begin{minipage}{0.235\textwidth}
        \centering \includegraphics[width=\linewidth]{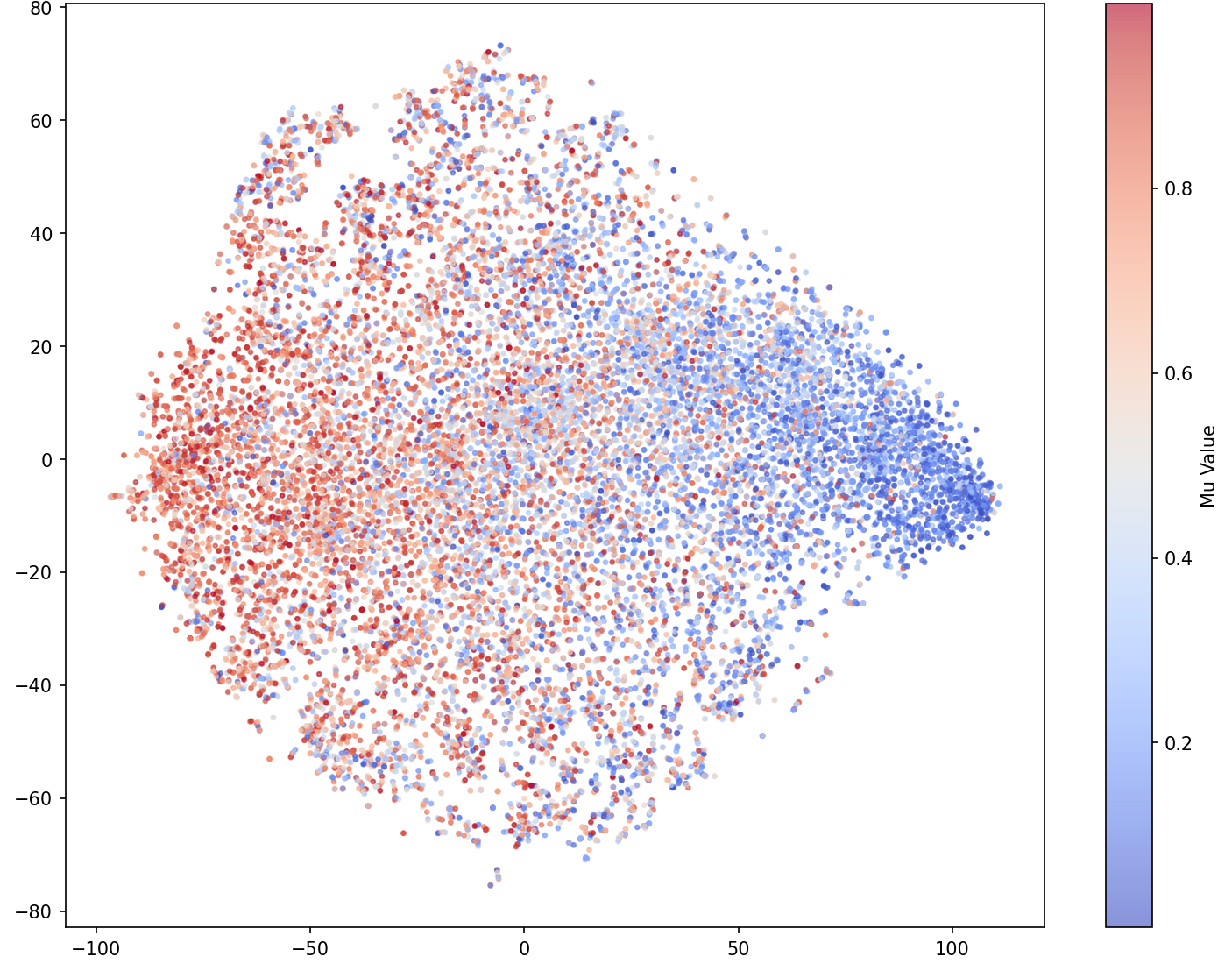}
        \centerline{(h) Full Model (PRISM)}
    \end{minipage}
    \vspace*{0mm}
    \caption{Latent embedding structure visualized by t-SNE. Embeddings are colored by ground-truth social stiffness $\mu$ ($\mu\approx0$: aggressive; $\mu\approx1$: cautious). RnC-based configurations show a clearer ordinal gradient than reconstruction- or regression-only variants.}
    \label{fig:tsne_ablation}
    \vspace*{-5mm}
\end{figure}

Table~\ref{tab:ablation} shows the quantitative trade-off among three requirements for an interaction-style representation. It should predict the simulated stiffness value, preserve motion information, and organize the latent space according to stiffness ordering. Manifold Rank $\rho$ is the Spearman rank correlation (denoted by $\operatorname{corr}_{\mathrm{S}}$) between all pairwise embedding distances and the corresponding pairwise stiffness differences:
$\rho=\operatorname{corr}_{\mathrm{S}}\!\left(\{\|\bm{z}_i-\bm{z}_j\|_2\}_{i<j},\{|\mu_i-\mu_j|\}_{i<j}\right)$.
Figure~\ref{fig:tsne_ablation} provides the complementary geometric view of the same trade-off using
%t-distributed stochastic neighbor embedding (t-SNE)
t-SNE and coloring each point by the ground-truth ORCA stiffness $\mu$ between 0 (blue) and 1 (red).

%%%The main observation from Fig.~\ref{fig:tsne_ablation} is not the appearance of any single panel. The progression from panels without ordinal supervision to panels with RnC. 
Figure~\ref{fig:tsne_ablation} highlights the effect of RnC supervision.
The T4P~\cite{Park2024T4P} representation in (a), Recon Only in (b), Reg Only in (c), and Recon + Reg in (e) are weakly organized by color: blue and red samples remain broadly mixed. Table~\ref{tab:ablation} gives the same conclusion quantitatively because their Manifold Rank values remain close to zero ($\rho=0.019$, $0.038$, and $0.043$ for Recon Only, Reg Only, and Recon + Reg). Thus, accurate reconstruction or scalar regression alone does not create a latent geometry that reflects relative behavioral similarity. This is important because the policy needs a latent state structured by interaction-style similarity.

The panels that include RnC (i.e., Fig.~\ref{fig:tsne_ablation} (d), (f), (g), and (h)) change this structure. They show a more coherent blue-to-red gradient. Table~\ref{tab:ablation} reports a corresponding increase in Manifold Rank ($\rho > 0.16$). This comparison indicates the role of RnC. It supplies the ordinal behavioral structure that is absent from reconstruction-only and regression-only objectives. However, RnC alone is not sufficient for navigation. Figures~\ref{fig:tsne_ablation} (d) and (f) show strong stiffness ordering, whereas Table~\ref{tab:ablation} shows poor Average Displacement Error (ADE)~\cite{gupta2018socialgan} for RnC Only and RnC + Reg. They capture the stiffness ordering of simulated agents, while losing much of the motion information needed for trajectory grounding.

The effectiveness of PRISM follows from the balance in Fig.~\ref{fig:tsne_ablation} (h). Compared with Recon + RnC in (g), the full model adds regression supervision and obtains the best MAE of stiffness (0.203). Compared with RnC + Reg in (f), it restores motion grounding, reducing reconstruction ADE from 0.771m to 0.027m. Compared with Recon + Reg in (e), it maintains an ordinal latent layout, raising Manifold Rank from 0.043 to 0.183. These linked comparisons show why all three objectives are used. RnC organizes the space by interaction style, reconstruction ties the style vector to observable trajectories, and regression stabilizes the stiffness estimate. The full PRISM representation is therefore suitable as a compact state variable for downstream social navigation.

\subsection{Results: Navigation Performance}

\begin{table}[t]
    \centering
    \caption{Navigation results under different human and obstacle densities.}
    \vspace*{-2mm}
    \label{tab:results_comparison}
    \resizebox{\textwidth}{!}{%
    \begin{tabular}{c l c c c c c c c}
        \toprule
        \multirow{2}{*}{\textbf{Environment}} & 
        \multirow{2}{*}{\textbf{Method}} & 
        \multirow{2}{*}{\textbf{Success}$\uparrow$} & 
        \multicolumn{3}{c}{\textbf{Collision Rate}$\downarrow$} & 
        \multirow{2}{*}{\textbf{Timeout}$\downarrow$} & 
        \multirow{2}{*}{\textbf{Nav Time (s)}$\downarrow$} & 
        \multirow{2}{*}{\textbf{Path Len (m)}$\downarrow$} \\
        \cmidrule(lr){4-6}
         & & & \textbf{Overall} & \textbf{w/ Humans} & \textbf{w/ Obs} & & & \\
        \midrule
        
        % Setting 1: Training Distribution (Default)
        \multirow{2}{*}{\shortstack[c]{\textbf{Training distribution}\\(5-9 humans, 8-12 obs.)}} 
          & HEIGHT (Baseline) & 0.81 & 0.19 & 0.13 & 0.06 & 0.00 & 18.78 & 10.30 \\
          & \textbf{PRISM (Ours)} & \textbf{0.88} & \textbf{0.12} & \textbf{0.08} & \textbf{0.04} & \textbf{0.00} & \textbf{18.08} & \textbf{10.20} \\
        \midrule

        % Setting 2: Less Crowded (Less Human)
        \multirow{2}{*}{\shortstack[c]{\textbf{Less crowded}\\(0-4 humans, 8-12 obs.)}} 
          & HEIGHT (Baseline) & 0.89 & 0.11 & 0.06 & 0.05 & 0.00 & 17.82 & 10.42 \\
          & \textbf{PRISM (Ours)} & \textbf{0.95} & \textbf{0.05} & \textbf{0.02} & \textbf{0.03} & \textbf{0.00} & \textbf{17.46} & \textbf{10.37} \\
        \midrule

        % Setting 3: More Crowded (More Human)
        \multirow{2}{*}{\shortstack[c]{\textbf{More crowded}\\(10-14 humans, 8-12 obs.)}} 
          & HEIGHT (Baseline) & 0.74 & 0.26 & 0.20 & 0.06 & 0.00 & 19.29 & 10.18 \\
          & \textbf{PRISM (Ours)} & \textbf{0.80} & \textbf{0.20} & \textbf{0.16} & \textbf{0.04} & \textbf{0.00} & \textbf{18.93} & \textbf{10.20} \\
        \midrule

        % Setting 4: Less Constrained (Less Obs)
        \multirow{2}{*}{\shortstack[c]{\textbf{Less constrained}\\(5-9 humans, 3-7 obs.)}} 
          & HEIGHT (Baseline) & 0.87 & 0.13 & 0.10 & 0.02 & 0.00 & 18.04 & 10.59 \\
          & \textbf{PRISM (Ours)} & \textbf{0.89} & \textbf{0.11} & \textbf{0.09} & \textbf{0.01} & \textbf{0.00} & \textbf{17.79} & \textbf{10.43} \\
        \midrule

        % Setting 5: More Constrained (More Obs)
        \multirow{2}{*}{\shortstack[c]{\textbf{More constrained}\\(5-9 humans, 13-17 obs.)}} 
          & HEIGHT (Baseline) & 0.73 & 0.27 & 0.11 & 0.16 & 0.01 & 19.30 & 10.41 \\
          & \textbf{PRISM (Ours)} & \textbf{0.77} & \textbf{0.23} & \textbf{0.08} & \textbf{0.15} & \textbf{0.00} & \textbf{18.85} & \textbf{10.26} \\
        
        \bottomrule
    \end{tabular}
    }
    \vspace*{-2mm}
\end{table}

Table~\ref{tab:results_comparison} evaluates whether the representation in Table~\ref{tab:ablation} and Fig.~\ref{fig:tsne_ablation} improves closed-loop navigation when inserted into HEIGHT.
%%%The comparison is deliberately strict: the baseline receives the same geometric observations, whereas PRISM additionally receives the inferred Interaction Style Vector and its confidence score. Therefore, improvements in Table~\ref{tab:results_comparison} indicate the value of behavior-aware state augmentation beyond geometry-only navigation.

Across all five evaluation regimes, PRISM reduces the overall collision rate. In the training-distribution setting, the collision rate decreases from 0.19 to 0.12, a relative reduction of 37\%, while success increases from 0.81 to 0.88. This consistency suggests that the inferred social stiffness variable helps the robot adjust its behavior around agents with different yielding tendencies.

The obstacle-density tests show a similar pattern, though the gain is smaller
under dense obstacles, as geometric bottlenecks cannot be fully resolved by
interaction-style reasoning alone. Nevertheless, PRISM consistently improves or maintains performance across regimes where social and geometric constraints are entangled.

Efficiency metrics show a modest secondary benefit. Navigation time is lower across all settings. Path length is also lower in four of the five settings, likely because better anticipation of yielding or assertive pedestrians reduces late evasive actions. Since Table~\ref{tab:results_comparison} reports aggregate rates and means without confidence intervals, these results should be viewed as empirical evidence in ORCA simulations rather than as a statistical significance claim. Even with this conservative interpretation, the results support the central claim that passively inferred heterogeneous interaction traits can improve reinforcement-learning-based robot navigation in dynamic crowds.

%%%%%%%%%%%%%%%%%%%%%%%%%%%%%%%%%%%%%%%%%%%%%%%%%%

\section{Discussion}

PRISM treats crowds as dynamic multi-agent worlds in which pedestrians are interactive agents rather than passive obstacles. PRISM infers interaction styles from passively observed motion histories and uses them as state variables for closed-loop robot planning. This formulation is closely aligned with interactive-agent and embodied-AI settings, where an agent must reason about how other agents respond during interaction.

In the ORCA-based simulator, PRISM reduces collision rates across all five evaluation regimes and yields small changes in navigation time and path length. These results suggest that interaction-style augmentation can improve navigation beyond geometric observations alone. Since personal-space violations are penalized independently of the inferred style, the policy is not explicitly rewarded for exploiting pedestrians who tend to yield.

The current evidence is limited to synthetic heterogeneity defined by an ORCA parameter, whereas real behavior is context-dependent and multidimensional. The results also lack confidence intervals, independent PPO training seeds, and a closed-loop ablation of its components, including an oracle-stiffness control. PRISM should therefore be viewed as a promising simulated interaction-representation module. Future work will evaluate it on real pedestrian data and learn multidimensional traits from naturally occurring interactions.

%%%%%%%%%%%%%%%%%%%%%%%%%%%%%%%%%%%%%%%%%%%%%%%%%%

\section{Conclusion}

This paper presented PRISM, a framework for behavior-aware robot navigation through passive inference of interaction styles. PRISM uses Rank-N-Contrast loss to organize trajectory embeddings by a simulated social stiffness parameter and injects the inferred representation into a robot navigation planner. Table~\ref{tab:ablation} and Fig.~\ref{fig:tsne_ablation} show a latent representation that is both ordered by interaction style and grounded in observed motion, and Table~\ref{tab:results_comparison} shows improved collision-avoidance metrics over a geometry-only baseline. The present results do not solve the sim-to-real problem of obtaining stiffness labels for real humans. Within controlled ORCA-based crowd simulations, however, PRISM demonstrates that passive latent-trait inference can support reinforcement-learning-based social navigation.
% PRISM is therefore positioned as a specialized motion-based representation module that can complement visual and contextual perception components, rather than as a complete model of human social behavior.

% ---- Bibliography ----

\bibliographystyle{splncs04}
\bibliography{main_aiw}
\end{document}